\pdfoutput=1

\documentclass[11pt]{article}

\usepackage[final]{acl}

\usepackage{times}
\usepackage{latexsym}

\usepackage[T1]{fontenc}

\usepackage[utf8]{inputenc}

\usepackage{microtype}

\usepackage{inconsolata}

\usepackage{graphicx}
\usepackage{enumitem}
\usepackage{multirow}
\usepackage{multicol}
\usepackage{booktabs} 
\usepackage{algorithm}       
\usepackage{algpseudocode}   
\usepackage{amsmath}         
\usepackage{amssymb}         
\usepackage{array}           
\usepackage{float}           
\usepackage{makecell}
\usepackage{url}
\usepackage{tcolorbox}
\usepackage{longtable}
\usepackage[table]{xcolor}
\usepackage[normalem]{ulem}
\tcbuselibrary{breakable}
\definecolor{navyblue}{RGB}{0,0,128}  
\title{GraphCheck: Multipath Fact-Checking with Entity-Relationship Graphs\thanks{Code available at: \url{https://github.com/windowh1/graphcheck}}}

\author{
\textbf{Hyewon Jeon},
\textbf{Jay-Yoon Lee} \\
Seoul National University\\
\texttt{\{pingpong0926,lee.jayyoon\}@snu.ac.kr}}

\begin{document}
\maketitle
\begin{abstract}
Automated fact-checking aims to assess the truthfulness of textual claims based on relevant evidence. However, verifying complex claims that require multi-hop reasoning remains a significant challenge. We propose \textbf{GraphCheck}, a novel framework that transforms claims into entity-relationship graphs for structured and systematic fact-checking. By explicitly modeling both explicit and latent entities and exploring multiple reasoning paths, GraphCheck enhances verification robustness. While GraphCheck excels in complex scenarios, it may be unnecessarily elaborate for simpler claims. To address this, we introduce \textbf{DP-GraphCheck}, a variant that employs a lightweight strategy selector to choose between direct prompting and GraphCheck adaptively. This selective mechanism improves both accuracy and efficiency by applying the appropriate level of reasoning to each claim. Experiments on the HOVER and EX-FEVER datasets demonstrate that our approach outperforms existing methods in verification accuracy, while achieving strong computational efficiency despite its multipath exploration. Moreover, the strategy selection mechanism in DP-GraphCheck generalizes well to other fact-checking pipelines, highlighting the broad applicability of our framework. 

\end{abstract}

\section{Introduction}
Automated fact-checking is a task that assesses the truthfulness of claims based on relevant evidence. With a standard pipeline that includes claim detection, evidence retrieval, and veracity assessment, automated systems enhance efficiency and accuracy in fact-checking~\cite{guo-etal-2022-survey}. However, verifying complex claims that require multi-hop reasoning remains a significant challenge. Such claims often consist of interwoven subclaims, making them difficult to verify at once. Also, relevant evidence is likely dispersed across multiple documents, complicating the retrieval process~\citep{jiang-etal-2020-hover, ma-etal-2024-ex}. 

Another key obstacle is the existence of latent entities—references not explicitly stated in the text. For example, in the claim \textit{``The musician, who is part of Tall Birds, is a percussionist for a band that formed in Issaquah, Washington''}, the phrases \textit{``The musician''} and \textit{``a band''} are latent entities. While these phrases correspond to specific entities (\textit{Davey Brozowski} and \textit{Modest Mouse}, respectively), the claim only implies relationships to these entities without explicitly revealing them. Identifying latent entities is crucial, as they can provide pivotal information for evidence retrieval and claim verification. Also, the order of latent entity identification matters, as some entities can be inferred more easily due to stronger contextual clues, which in turn help identify others as well. Conversely, initially misidentifying a challenging entity introduces false information, hindering subsequent identification steps. Thus, latent entity identification plays an important role in fact-checking.

\begin{figure*}[t!]
    \includegraphics[width=\textwidth]{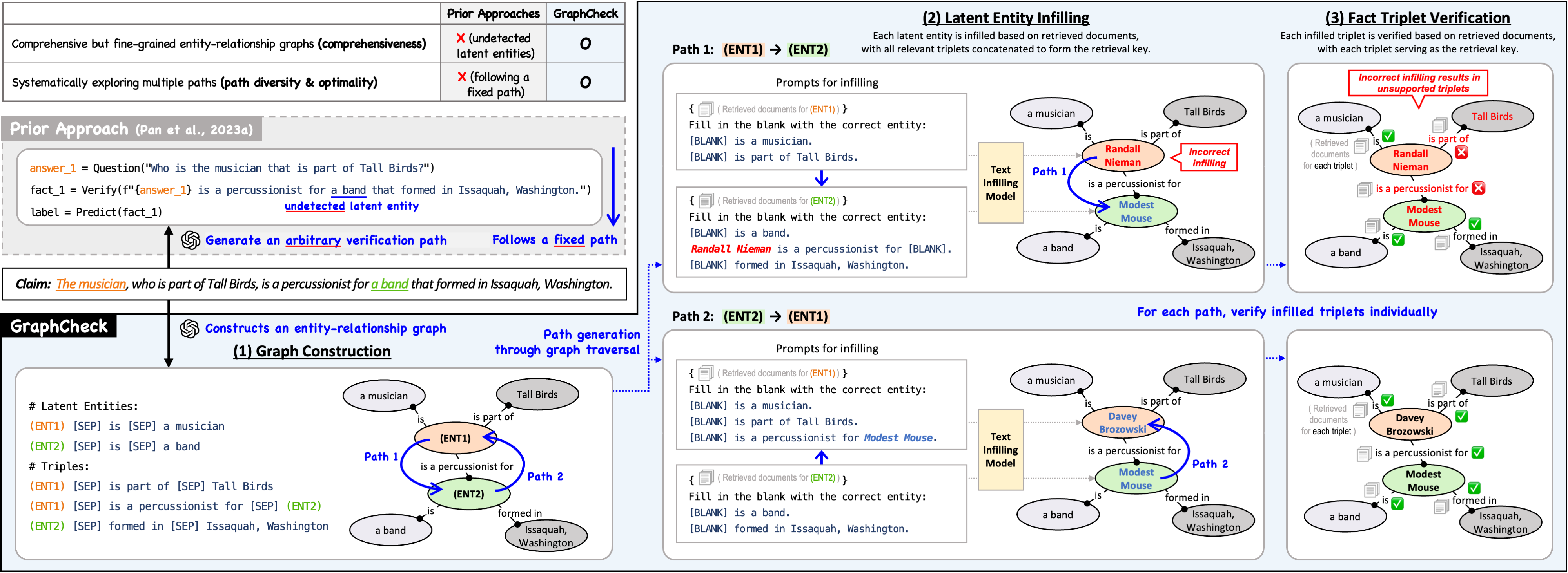}
    \caption{Overview of GraphCheck. Compared to prior approaches, GraphCheck offers a comprehensive yet fine-grained claim decomposition. It also systematically explores multiple paths instead of following a fixed path. The overall process consists of three steps: (1) A claim is converted into a structured entity-relationship graph in which both explicit and latent entities are represented. (2) Latent entities are then identified through text infilling. Multiple infilling paths are explored via graph traversal, resulting in multiple infilled graphs. (3) For each infilled graph, every triplet is individually verified. The claim is predicted as \textsc{Supported} if at least one path yields a graph in which all triplets are verified as \textsc{Supported}; otherwise, the claim is predicted as \textsc{Not Supported}.}
    \label{fig:overview}
    \vspace{-8pt}
\end{figure*}

Recent work has examined the application of large language models (LLMs) for verifying multi-hop claims, relying solely on few-shot prompting without additional task-specific training~\citep{brown2020languagemodelsfewshotlearners, dmonte2024claimverificationagelarge}. This approach is appealing due to its scalability and generality, yet it still faces notable limitations. Prior methods leverage LLMs to generate verification paths by identifying ``check-worthy'' components~\cite{guo-etal-2022-survey}, such as subclaims to be verified and question-answering steps for latent entity identification~\citep{pan-etal-2023-fact, pan-etal-2023-qacheck, wang2023explainable}. However, due to the inherent ambiguity of check-worthiness, verification paths often lack granularity or omit key components. Moreover, these methods typically rely on an LLM-generated, fixed verification path that may not be optimal, ultimately limiting verification accuracy.

To address these limitations, we propose \textbf{GraphCheck}, in which LLMs transform text-based claims into structured entity-relationship graphs through which diverse verification paths can be generated. These graphs consist of fact triplets, each defining a relationship between entities and serving as an independently verifiable subclaim. Compared to prior approaches that extract check-worthy components without a clear structure, GraphCheck effectively performs fine-grained claim decomposition while preserving key components, enabling more comprehensive verification. Furthermore, by enabling flexible graph traversal, GraphCheck avoids reliance on a single reasoning path and allows multiple orders of latent entity identification. This flexibility increases the likelihood of capturing an optimal reasoning path, improving the robustness of verification.

While GraphCheck thoroughly verifies complex claims, it may be unnecessarily elaborate for simpler cases. Some relatively simple claims can be verified more efficiently and effectively through direct prompting (\textbf{DP} or \textbf{Direct}), where the LLM directly assesses a claim’s truthfulness based on relevant documents. To leverage the complementary strengths of Direct and GraphCheck, we introduce \textbf{DP-GraphCheck}, which employs a lightweight \emph{strategy selector} to choose between the two methods adaptively. Notably, this strategy selector is modular and can be easily integrated into diverse fact-checking systems beyond GraphCheck.

Experimental results on multi-hop fact-checking datasets~\citep{jiang-etal-2020-hover, ma-etal-2024-ex} demonstrate that both GraphCheck and DP-GraphCheck outperform existing methods on the complex multi-hop fact-checking task, highlighting the effectiveness of structuring claims as entity-relationship graphs. Despite its multipath exploration, GraphCheck achieves superior computational efficiency compared to multipath baselines. Moreover, the strategy selector improves performance not only within GraphCheck but also across other baseline fact-checking systems, underscoring its broad applicability.

\section{Related Work}
\paragraph{Multi-hop Fact-Checking}

Fact-checking involves assessing the veracity of claims based on supporting evidence. Early research primarily focuses on \emph{single-hop fact-checking}, where the evidence necessary to validate a claim is contained within a single document or passage~\citep{vlachos-riedel-2014-fact, wang-2017-liar, thorne-etal-2018-fever}. To better reflect real-world situations, where claim verification often depends on dispersed or interconnected information, subsequent work proposes \emph{multi-hop fact-checking}, which demands reasoning across multiple pieces of evidence~\citep{jiang-etal-2020-hover, ma-etal-2024-ex, aly-etal-2021-fact}.

Early approaches to multi-hop fact-checking rely on task-specific supervised training of neural models with annotated datasets~\citep{jiang-etal-2020-hover, ijcai2021p536, khattab2022baleenrobustmultihopreasoning}. While these methods demonstrate strong in-domain performance and maintain computational efficiency during inference through lightweight architectures, they are fundamentally limited by the high cost of large-scale annotation and poor cross-domain generalization.

The advent of large language models (LLMs) has shifted the field toward more flexible, context-aware reasoning~\citep{guo-etal-2022-survey,dmonte2024claimverificationagelarge}. 
LLM-based approaches often combine retrieval-augmented generation (RAG)~\citep{lewis2021retrievalaugmentedgenerationknowledgeintensivenlp,gao2024retrievalaugmentedgenerationlargelanguage} with in-context learning~\cite{brown2020languagemodelsfewshotlearners} to support veracity prediction without additional task-specific training. 
Recent work has explored various strategies, such as iterative RAG~\cite{shao-etal-2023-enhancing} and claim decomposition~\citep{chen2023combatingmisinformationagellms,zhang-gao-2023-towards,pan-etal-2023-fact}. 
Building on these advances, GraphCheck introduces a graph-based decomposition approach that structures claims into entities and relations, enabling more comprehensive multi-hop fact-checking.


\paragraph{Latent Entity Identification}

Identifying latent entities is crucial for improving evidence retrieval and verification accuracy in multi-hop fact-checking. Previous approaches have addressed this challenge using question-answering frameworks~\citep{pan-etal-2023-qacheck, pan-etal-2023-fact, wang2023explainable}. GraphCheck takes a different approach by constructing an entity-relationship graph where latent entities are represented as placeholder nodes and identified through text infilling~\cite{zhu2019textinfilling}.

Regarding the ordering of latent entity identification, existing methods follow model-driven approaches with distinct strategies. Specifically, ~\citet{shao-etal-2023-enhancing} and ~\citet{pan-etal-2023-qacheck} employ iterative processes where each step builds on previous outputs, creating linear reasoning chains, while ~\citet{pan-etal-2023-fact} generates a complete reasoning path within a single LLM call and aggregates results across multiple generated paths. In contrast, GraphCheck systematically explores multiple identification paths within a graph structure constructed through a single API call, thereby achieving both efficiency and robustness.


\paragraph{Graph-based Fact-Checking}

Recent advances in fact-checking have increasingly adopted graph-based methods to represent structured relationships in textual claims or evidence. GEAR~\cite{zhou-etal-2019-gear} models each evidence sentence as a node and applies graph neural networks to aggregate information across sentences. KGAT~\cite{liu-etal-2020-fine} represents claim–evidence pairs as nodes and applies kernel-based attention over the graph.

Other approaches focus on entity-level relationships. Zhong et al.~\cite{zhong-etal-2020-reasoning} employ semantic role labeling to construct graphs from evidence, while Yuan and Vlachos~\cite{yuan-vlachos-2024-zero} extract triplets from claims with OpenIE models and verify them using NLI models. GraphCheck takes a distinct approach by constructing entity–relationship graphs that represent latent entities as placeholder nodes, leveraging the in-context learning capabilities of LLMs.

\section{Methodology}
\subsection{GraphCheck: Graph Construction}

GraphCheck first transforms a claim into an entity-relationship graph. Latent entities within the claim are detected and represented as placeholder nodes. To construct the graph, we leverage an LLM guided by instructions that include predefined rules and few-shot examples (Appendix~\ref{appendix:prompt_graph_construction}). The key instructions are as follows:

\begin{itemize}[itemsep=1pt, parsep=0pt, topsep=1pt]
    \item Detect and represent latent entities using placeholders (e.g., \texttt{\small(ENT1)}, \texttt{\small(ENT2)}).
    \item Decompose the claim into fact triplets (e.g., \texttt{\small subject [SEP] relation [SEP] object}), each serving as a basic unit of information.
\end{itemize}

As shown in the part of Figure \ref{fig:overview} that explains the \emph{(1) Graph Construction} step of GraphCheck, the generated graph consists of two sections:

\begin{itemize}[itemsep=1pt, parsep=0pt, topsep=1pt]
    \item \emph{\# Latent Entities}: Triplets that link latent entities to their implicit references in the claim.
    \item \emph{\# Triples}: Triplets that capture relationships between entities.
\end{itemize}

While triplets in both sections serve as subclaims requiring verification, the \emph{\# Latent Entities} section is set apart to ensure that contextual meaning is preserved when placeholders are introduced. For instance, in the example shown in Figure \ref{fig:overview}, replacing \textit{``the musician''} with a placeholder \texttt{\small(ENT1)} could result in a loss of information. To prevent this, placeholders are explicitly mapped to their corresponding references in the separated section.


\subsection{GraphCheck: Latent Entity Infilling}

\paragraph{Latent Entity Identification}

Once the entity-relationship graph is constructed, latent entities are identified sequentially through text infilling. The process begins by retrieving top-$k$ documents relevant to the target latent entity. To ensure sufficient context, a retrieval query is formulated by concatenating all triplets that include the target entity and exclude any other unidentified latent entities. The retrieved documents, along with the same set of triplets, are then used as input to the text infilling model. A more detailed description of the infilling process is provided in Appendix~\ref{appendix:prompt_infilling}.

The part of Figure \ref{fig:overview} that explains the \emph{(2) Latent Entity Infilling} step shows an example of the infilling process. In Path 1, \texttt{\small(ENT1)} is identified first as the target entity. Then, \texttt{\small(ENT2)} becomes the next target entity, at which point \texttt{\small(ENT1)} has already been identified as \textit{``Randall Nieman''}, providing additional contextual information.


\paragraph{Multipath Exploration} 

For claims containing multiple latent entities, various identification orders are explored. In Figure \ref{fig:overview}, two possible paths are considered: \texttt{\small(ENT1)}{\small$\rightarrow$}\texttt{\small(ENT2)} and \texttt{\small(ENT2)}{\small$\rightarrow$}\texttt{\small(ENT1)}. To manage computational complexity, up to $\bar{P}$ paths are randomly sampled if the total number of possible orders exceeds this limit.

Exploring various identification paths is important because some paths can be more effective than others. Some latent entities are identified more easily due to stronger contextual clues, and identifying them first can provide additional context for subsequent identifications. For example, in Figure \ref{fig:overview}, identifying \texttt{\small(ENT2)} is easier than identifying \texttt{\small(ENT1)} because \textit{``Issaquah, Washington''} provides a more salient retrieval cue than \textit{``Tall Birds''}, which may introduce ambiguity between a rock band and actual birds. As a result, when \texttt{\small(ENT2)} is identified first in \emph{Path 2}, both entities are correctly infilled, whereas \emph{Path 1} fails to infill \texttt{\small(ENT1)} correctly.

However, automatically finding the optimal path is challenging. To address this, our method systematically explores multiple paths instead of relying on model-driven planning, thereby increasing the likelihood of finding the most effective path.


\subsection{GraphCheck: Fact Triplet Verification}

\paragraph{Triplet Verification} 

After infilling, each triplet in the graph is independently verified. For each infilled triplet $t'$, the top-$k$ documents are retrieved from the corpus using $t'$ as the retrieval query. The verifier assesses the veracity of $t'$ using $k{+}1$ evidence inputs: (i) the concatenation of the top-$k$ documents, and (ii) each of the top-$k$ documents individually. If any of these inputs yields a \textsc{Supported} judgment, $t'$ is classified as \textsc{Supported}; otherwise, it is classified as \textsc{Not Supported}.


\paragraph{Path Verification} 

A latent entity identification path produces a fully infilled graph consisting of triplets $\{t_1', t_2', \dots, t_n'\}$. The path is classified as \textsc{Supported} if all triplets in the graph are classified as \textsc{Supported}; otherwise, it is classified as \textsc{Not Supported}.


\paragraph{Claim Verification} 

Since multiple identification paths can be explored, GraphCheck performs triplet-level verification for each path independently. A claim is \emph{ultimately} classified as \textsc{Supported} if at least one path is classified as \textsc{Supported}; otherwise, it is \emph{ultimately} classified as \textsc{Not Supported}.

This classification approach accounts for potential errors in latent entity infilling. A claim classified as \textsc{Not Supported} in a single path does not necessarily indicate that the claim itself is false; rather, it may stem from incorrectly identified latent entities. By exploring multiple identification paths, GraphCheck increases the likelihood of accurate verification, as the claim veracity can be reliably assessed if at least one path correctly identifies latent entities.


\subsection{DP-GraphCheck}

GraphCheck rigorously evaluates the supportedness of a claim; a claim is classified as \textsc{Supported} only if the infilling results of all triplets align with the retrieved evidence. However, for relatively simple claims that do not require decomposition or latent entity identification, this approach can be unnecessarily strict and may misclassify correct claims as \textsc{Not Supported}.

To address this limitation, we introduce \textbf{DP-GraphCheck}, which improves both efficiency and accuracy of GraphCheck. Given the claim and its top-$k$ retrieved documents (using the original claim itself as a query), a lightweight \textit{strategy selector} determines whether the retrieved evidence is sufficient for assessing its veracity. If deemed sufficient, the claim is considered simple and verified with Direct. Otherwise, the claim undergoes the full GraphCheck pipeline, including claim decomposition and latent entity infilling.

In short, DP-GraphCheck efficiently filters out simpler claims while maintaining thorough verification for more complex cases. The complete verification process of this framework is summarized in Appendix~\ref{appendix:algorithm}.

\section{Experiments}
\subsection{Experimental Setup}

\paragraph{Datasets} 

We utilize two datasets for evaluation:

\begin{itemize}[itemsep=1pt, parsep=0pt, topsep=1pt]

    \item \textbf{HOVER}~\cite{jiang-etal-2020-hover} is a dataset for multi-hop fact-checking, verifying whether a claim is supported or not based on evidence dispersed across multiple Wikipedia articles (2 to 4 hops). Since the test set labels are not publicly available, we use the development set as our test set. We utilize the preprocessed October 2017 Wikipedia dump~\cite{yang-etal-2018-hotpotqa} as the retrieval corpus.

    \item \textbf{EX-FEVER}~\cite{ma-etal-2024-ex} is another multi-hop fact-checking dataset where evidence is scattered across multiple Wikipedia articles (2 to 3 hops). Unlike HOVER, which has only two labels, EX-FEVER introduces an additional ``Not Enough Information (NEI)'' label. We exclude NEI-labeled samples, as the label does not necessarily indicate the absence of evidence in the entire retrieval corpus, but only in the annotated subset. We use the preprocessed Wikipedia dump provided by \citet{jiang-etal-2020-hover} as the retrieval corpus.

\end{itemize}


\paragraph{Implementation Details}
Our framework employs \texttt{\small flan-t5-xl}~\cite{chung2022scalinginstructionfinetunedlanguagemodels} for text infilling, fact triplet verification, strategy selection, and Direct, applying task-specific prompts (Appx.\ref{appendix:prompts}) with greedy decoding. We use the Hugging Face checkpoint without additional task-specific training. This setup aligns with our baseline~\cite{pan-etal-2023-fact}, ensuring a fair comparison.

For document retrieval, we adopt BM25~\cite{10.1561/1500000019}, maintaining consistency with \citet{pan-etal-2023-fact} as well. Our primary focus is the \emph{open-book} setting, where the verification is conducted based on the top-$k$ retrieved documents ($k{=}10$). We also evaluate performance in the \emph{open-book + gold} setting, where the claim's gold document set is merged with the top retrieved documents to form a set of $k$ documents.

We employ \texttt{\small gpt-4o-2024-08-06} for graph construction with temperature=0.0 and top\_p=1.0. The few-shot examples used in the prompt are manually annotated using 10 instances randomly sampled from the HOVER training set (Appx.\ref{appendix:prompt_graph_construction}). 

The path limit $\bar{P}$, which defines the maximum number of exploration paths, is set to 5. Additionally, $\bar{P}{=}1$ is also tested to assess the impact of multiple path exploration.


\paragraph{Baselines}
We compare our approach against several fact-checking frameworks that rely on in-context learning:

\begin{itemize}[itemsep=1pt, parsep=0pt, topsep=1pt]

    \item \textbf{ProgramFC}~\cite{pan-etal-2023-fact} converts complex claims into Python-like reasoning programs, outlining step-by-step actions such as question answering and subclaim verification. We generally follow the original setup; however, since ProgramFC originally uses \texttt{\small Codex}~\cite{chen2021evaluatinglargelanguagemodels}, we re-run the experiments using \texttt{\small  gpt-4o} for comparability. We evaluate two cases, $N{=}1$ and $N{=}5$, where $N$ represents the number of LLM API calls, each using stochastic decoding to produce a distinct reasoning program.

    \item \textbf{FOLK}~\cite{wang2023explainable} decomposes claims into First-Order Logic (FOL) clauses and question-answering sets required for claim verification. Originally, FOLK uses \texttt{\small gpt-3.5} (\texttt{\small text-davinci-003}) for decomposing and SerpAPI for evidence retrieval and question answering. To align with our setup, we adapt FOLK to use \texttt{\small  gpt-4o} and replace its question-answering module with \texttt{\small flan-t5-xl}, which generates answers based on Wikipedia articles retrieved via BM25.
    
    \item \textbf{Direct (DP)}~\cite{chung2022scalinginstructionfinetunedlanguagemodels} prompts the LLM to verify a claim based on documents retrieved using the original claim as the query (Appx.\ref{tab:prompt_direct}). We implement it using \texttt{\small flan-t5-xl}. Since it also serves as the Direct component of DP-GraphCheck, we report its standalone performance as well.

    \item We apply our proposed strategy selector to ProgramFC and FOLK, denoted as \textbf{DP-ProgramFC} and \textbf{DP-FOLK}. These variants serve as fair baselines for comparison with DP-GraphCheck and help assess the strategy selector’s generalizability across different fact-checking frameworks.
    
\end{itemize}

Although our main evaluation focuses on in-context learning baselines, we also perform comprehensive comparisons with fine-tuned fact-checking models in Appx.\ref{appendix:ft_model}, where GraphCheck consistently achieves superior performance—particularly in realistic settings that involve retrieval noise and domain shift.


\begin{table*}[t!]
\small
\centering
\renewcommand{\arraystretch}{1.3}
\resizebox{\textwidth}{!}{
\begin{tabular}{l | ccc | cc | ccc | cc | c | c}
    \toprule
    \multirowcell{3}[-6pt][l]{\textbf{Methods}} 
    & \multicolumn{5}{c|}{\textbf{Open-book}} 
    & \multicolumn{5}{c|}{\textbf{Open-book + Gold}} 
    & \multirowcell{4}[0pt][c]{\textbf{Average}\\\textbf{Runtime}\\[1pt]\scriptsize(minutes \\\scriptsize1k samples)} 
    & \multirowcell{4}[0pt][c]{\textbf{Average}\\\textbf{API Cost}\\[1pt]\scriptsize(USD per\\\scriptsize1k samples)} \\
    \cmidrule(lr){2-6} \cmidrule(lr){7-11} 
     & \multicolumn{3}{c|}{\textbf{HOVER}} 
     & \multicolumn{2}{c|}{\textbf{EX-FEVER}} 
     & \multicolumn{3}{c|}{\textbf{HOVER}} 
     & \multicolumn{2}{c|}{\textbf{EX-FEVER}} 
     & & \\
    \cmidrule(lr){2-4} \cmidrule(lr){5-6} \cmidrule(lr){7-9} \cmidrule(lr){10-11} 
     & \textbf{2-hop} & \textbf{3-hop} & \textbf{4-hop} 
     &  \textbf{2-hop} & \textbf{3-hop} 
     & \textbf{2-hop} & \textbf{3-hop} & \textbf{4-hop} 
     & \textbf{2-hop} & \textbf{3-hop} 
     & & \\
    \specialrule{0.8pt}{1pt}{3pt}
    Direct (DP) & 72.56 & 61.70 & 59.57 & 81.03 & 73.02 & 76.03 & 67.18 & 61.26 & \textbf{87.77} & 81.82 & 4.94 & 0.00 \\
    \midrule
    ProgramFC ($N{=}1$) & 70.04 & 61.33 & 59.00 & 77.55 & 71.50 & 71.52 & 64.74 & 63.99 & 83.82 & 78.52 & 49.83 & 3.39 \\
    DP-ProgramFC ($N{=}1$) & 70.79 & 62.75 & 60.61 & 79.46 & 74.62 & 71.19 & 66.04 & 65.31 & 83.42 & 78.93 & 37.63 & 2.12 \\
    ProgramFC ($N{=}5$) & 70.29 & 61.82 & 60.19 & 78.31 & 72.16 & 70.73 & 65.50 & 63.79 & 84.09 & 79.69 & 279.22 & 17.20 \\
    DP-ProgramFC ($N{=}5$) & 70.71 & 63.39 & 61.56 & 79.78 & \underline{75.30} & 70.52 & 66.23 & 64.86 & 83.69 & 79.80 & 180.97 & 10.76 \\
    FOLK & 65.13 & 59.63 & 56.10 & 72.32 & 63.85 & 69.96 & 66.23 & 65.65 & 80.13 & 75.98 & 111.78 & 7.71 \\
    DP-FOLK & 70.96 & 63.53 & 58.36 & 80.64 & 73.68 & 72.48 & 69.39 & 68.59 & 84.99 & 80.66 & 76.05 & 4.81 \\
    \midrule
    \textbf{GraphCheck ($\bar{P}{=}1$)} & 73.05 & 64.87 & 59.19 & 75.71 & 65.02 & \underline{78.18} & 70.68 & 67.70 & 83.06 & 75.86 & 73.46 & 3.00 \\    
    \textbf{DP-GraphCheck ($\bar{P}{=}1$)} & \textbf{76.29} & 67.36 & 62.35 & \textbf{81.12} & 74.56 & 77.25 & 73.00 & 71.95 & \underline{85.78} & \textbf{82.87} & 51.86 & 1.87 \\
    \midrule
    \textbf{GraphCheck ($\bar{P}{=}5$)} & 74.12 & \underline{67.71} & \underline{64.79} & 76.56 & 69.94 & \textbf{78.59} & \textbf{73.78} & \underline{72.55} & 83.64 & 80.16 & 88.05 & 3.00 \\ 
    \textbf{DP-GraphCheck ($\bar{P}{=}5$)} & \textbf{76.29} & \textbf{68.70} & \textbf{66.64} & \textbf{81.12} & \textbf{76.02} & 76.96 & \underline{73.34} & \textbf{73.63} & 85.69 & \underline{82.73} & 62.48 & 1.87 \\
    \bottomrule
\end{tabular}
}
\caption{Macro-F1 scores under open-book and open-book + gold settings, along with average runtime (minutes) and average API cost (USD) per 1k samples. The best Macro-F1 score in each column is highlighted in \textbf{bold}, and the second-best is \underline{underlined}. Note that Direct incurs no API cost, as it utilizes the open-source model. DP-GraphCheck ($\bar{P}{=}5$) outperforms most cases while being 2.9$\times$ faster and incurring 5.8$\times$ lower API cost than the best baseline.}

\label{tab:main_result}
\end{table*}

\subsection{Main Result}

Table~\ref{tab:main_result} summarizes the Macro-F1 scores for both the open-book and open-book + gold settings.


\paragraph{DP-GraphCheck}

In nearly all configurations—on the HOVER and EX-FEVER datasets, under both open-book and open-book + gold settings—DP-GraphCheck ($\bar{P}{=}5$) achieves either the best or second-best Macro-F1 score. 

An exception occurs on the 2-hop of the EX-FEVER dataset under the open-book + gold setting, where Direct achieves the highest score. This can be attributed to EX-FEVER's extractive nature, where claims closely match phrases in the gold documents with minimal rephrasing or abstraction (see Appx.\ref{appendix:ex-fever_example} for a specific example). In such cases, Direct, which preserves the claim's original form, may yield strong results. However, this advantage is less likely to generalize to real-world scenarios, where claims often diverge in wording from the supporting evidence. In these settings, structured reasoning becomes critical for reliable fact-checking.


\paragraph{GraphCheck}

On the HOVER dataset, under both the open-book and open-book + gold settings, GraphCheck ($\bar{P}{=}5$) outperforms all baselines even without the strategy selector. This highlights the effectiveness of graph-based structured verification in scenarios that require multi-hop reasoning.

On the EX-FEVER dataset, GraphCheck occasionally underperforms compared to baseline methods. Considering the extractive nature of EX-FEVER, these results may be attributed to GraphCheck's enforced fine-grained decomposition, in which the claim is broken down into a set of entity-relation triplets, each corresponding to a subclaim. While such structured decomposition is beneficial for complex reasoning, it may be unnecessary for extractive-style claims that can be verified holistically. However, when combined with Direct (i.e., DP-GraphCheck), GraphCheck consistently recovers strong performance.


\paragraph{Multipath Exploration}

We also observe that GraphCheck with $\bar{P}{=}5$ generally outperforms its $\bar{P}{=}1$ counterpart, and this improvement carries over to the DP-GraphCheck. The performance gap between multipath and single path variants becomes more pronounced as the hop count increases. For example, in the HOVER open-book setting, GraphCheck ($\bar{P}{=}5$) surpasses GraphCheck ($\bar{P}{=}1$) by 1.07, 2.84, and 5.60 points in 2-hop, 3-hop, and 4-hop claims, respectively. These results indicate that multipath exploration becomes increasingly beneficial for more complex claims.

While ProgramFC also attempts multipath verification by generating $N$ distinct verification programs via $N$ independent LLM API calls, the performance gap between ProgramFC ($N{=}1$) and ProgramFC ($N{=}5$) remains relatively small. This suggests that the reasoning paths generated by each call may lack sufficient diversity despite stochastic decoding. This method also incurs a high computational cost due to repeated LLM usage, even for claims that may not require extensive reasoning. In contrast, GraphCheck constructs a single entity-relationship graph through one LLM call, from which diverse paths can be explored. This design naturally supports reasoning diversity without additional LLM overhead.


\paragraph{Efficiency and Cost Analysis}
We evaluate runtime and API cost on an NVIDIA H100 and the OpenAI API (Table~\ref{tab:main_result}). The best-performing DP-GraphCheck ($\bar{P}{=}5$) achieves a 2.9$\times$ faster runtime and a 5.8$\times$ lower cost than the multipath baseline DP-ProgramFC ($N{=}5$). These results demonstrate that GraphCheck effectively balances verification robustness and efficiency, making it practical for real-world deployment.

Despite exploring multiple paths, GraphCheck achieves high efficiency through several design choices: (i) enforcing a path limit ($\bar{P}{=}5$) to cap the number of explorations, (ii) enabling multipath exploration without repeated LLM calls, and (iii) allocating the number of paths adaptively based on claim complexity. (iv) Furthermore, DP-GraphCheck improves the efficiency of GraphCheck by incorporating the lightweight Direct method through the strategy selector only when deemed appropriate.

As shown in Appx.\ref{appendix:num_hops_analysis}, higher-hop claims tend to involve more latent entities and thus allow more identification orders, whereas simpler claims involve fewer. Unlike prior methods that apply a fixed number of paths ($N$) to all claims, GraphCheck avoids unnecessary computation by tailoring the number of paths to each claim. 


\paragraph{Effectiveness of Strategy Selector}

The strategy selector proves effective across all baselines: both DP-ProgramFC and DP-FOLK consistently outperform their original counterparts. This demonstrates that the strategy selector enhances not only our framework, but also improves the performance of other fact-checking methods.


\subsection{Ablation Study}

\paragraph{Breakdown of DP-GraphCheck Performance}

Table~\ref{tab:dp_selector_breakdown} presents a detailed breakdown of DP-GraphCheck's performance on the HOVER dataset under the open-book setting. It reports results for the entire dataset and for two subsets of claims, each assigned to either Direct or GraphCheck by the strategy selector. For each group, the table shows the proportion of claims, the retrieval recall when querying with the original claim (as done by both the strategy selector and Direct), and the verification accuracy of both Direct and GraphCheck.

We observe that the strategy selector assigns an increasing proportion of claims to GraphCheck as hop count increases: 59.68\% of 2-hop claims are handled by GraphCheck, compared to 88.35\% for 4-hop claims. This trend indicates that the strategy selector is capable of discerning claim difficulty and assigning complex cases to the more systematic fact-checking method.

\begin{table}[t!]
    \centering
    \renewcommand{\arraystretch}{1.3}
    \resizebox{\columnwidth}{!}{
    \begin{tabular}{clccc}
    \toprule
    \textbf{Group} & \textbf{Metric} & \textbf{2-hop} & \textbf{3-hop} & \textbf{4-hop} \\
    \specialrule{0.8pt}{1pt}{1pt}
    \multirow{3}{*}{\makecell[c]{\textbf{Total}}} 
    & Recall@10              & 73.18  & 51.34  & 36.43 \\
    & Accuracy (Direct)      & 72.56  & 62.02  & 59.58 \\
    & Accuracy (GraphCheck)  & 74.60  & 68.12  & 66.79 \\
    \specialrule{0.8pt}{1pt}{1pt}
    \multirow{4}{*}{\makecell[c]{\textbf{Assigned}\\\textbf{to}\\\textbf{Direct}}}
    & \% of Claims                & 40.32\% & 23.22\% & 11.65\% \\
    & Recall@10                  & 84.47  & 64.32  & 55.79 \\
    & \textbf{Accuracy (Direct)} & \textbf{74.01}  & \textbf{71.13}  & \textbf{72.73} \\
    & Accuracy (GraphCheck)      & 69.82  & 68.54  & 64.46 \\
    \midrule
    \multirow{4}{*}{\makecell[c]{\textbf{Assigned}\\\textbf{to}\\\textbf{GraphCheck}}}
    & \% of Claims                   & 59.68\% & 76.78\% & 88.35\% \\
    & Recall@10                      & 65.55  & 47.41  & 33.88 \\
    & Accuracy (Direct)              & 71.58  & 59.26  & 57.84 \\
    & \textbf{Accuracy (GraphCheck)} & \textbf{77.83}  & \textbf{67.99}  & \textbf{67.10} \\
    \bottomrule
    \end{tabular}
    }
    \caption{Breakdown of DP-GraphCheck performance on the HOVER dataset under open-book setting. Results are grouped by the strategy selector's assignment: total samples, samples assigned to Direct, and those assigned to GraphCheck. Bold values indicate the accuracy of the fact-checking method applied in each group. The results show that the strategy selector assigns cases with enough evidence (high recall) to Direct and opposite cases to the GraphCheck module for a more fine-grained analysis. As a result, the combined model benefits from the best of both worlds. Similar trends can be seen in other baselines (Table~\ref{tab:main_result}).}
    \label{tab:dp_selector_breakdown}
\end{table}

Moreover, claims assigned to Direct exhibit consistently higher retrieval recall than those assigned to GraphCheck. This indicates that the strategy selector effectively assesses whether retrieved evidence is sufficient to support Direct verification, thereby avoiding unnecessary use of more complex fact-checking procedures.

In terms of accuracy, within the group of claims assigned to Direct, the accuracy of the Direct consistently exceeds that of GraphCheck across all hop levels, demonstrating the effectiveness of direct prompting for simpler claims. Conversely, for claims assigned to GraphCheck, the GraphCheck method outperforms Direct, highlighting the importance of structured reasoning in complex scenarios.

Overall, these results demonstrate that DP-GraphCheck successfully combines the strengths of Direct and GraphCheck through the strategy selector, achieving not only greater efficiency but also improved overall performance.

\begin{table}[t!]
    \centering
    \renewcommand{\arraystretch}{1.3}
    \resizebox{\columnwidth}{!}{
    \begin{tabular}{lllccc}
    \toprule
    \textbf{Backbone Model} & \textbf{Size} & \textbf{Method} & \textbf{2-hop} & \textbf{3-hop} & \textbf{4-hop} \\
    \midrule
    \rowcolor{gray!15}
    \multicolumn{6}{c}{\textbf{Prompting-Only Models}} \\
    \multirow{3}{*}{\texttt{gpt-4o \textbf{(Default)}}} 
    & \multirow{3}{*}{--} 
    & DP-ProgramFC       & 70.71 & 63.39 & 61.56 \\
    \cmidrule(lr){3-6}
    & & GraphCheck       & 74.12 & 67.71 & 64.79 \\
    & & DP-GraphCheck    & 76.29 & 68.70 & 66.64 \\
    \midrule
    \multirow{2}{*}{\texttt{claude-3-5-sonnet}} 
    & \multirow{2}{*}{--}
    & GraphCheck       & 76.39 & 68.53 & 63.65 \\
    & & DP-GraphCheck    & 76.45 & 69.74 & 65.50 \\
    \midrule
    \multirow{2}{*}{\texttt{gpt-3.5-turbo}} 
    & \multirow{2}{*}{--}
    & GraphCheck       & 70.06 & 59.93 & 59.35 \\
    & & DP-GraphCheck    & 74.42 & 63.68 & 62.00 \\
    \midrule
    \multirow{2}{*}{\texttt{Qwen2.5-72B-Instruct}} 
    & \multirow{2}{*}{72B}
    & GraphCheck       & 73.04 & 62.33 & 62.74 \\
    & & DP-GraphCheck    & 77.00 & 66.49 & 65.55 \\
    \midrule
    \rowcolor{gray!15}
    \multicolumn{6}{c}{\textbf{Fine-tuned Model}} \\
    \multirow{2}{*}{\texttt{flan-t5-xl}} 
    & \multirow{2}{*}{3B}
    & GraphCheck       & 74.77 & 66.55 & 61.37 \\
    & & DP-GraphCheck    & 76.64 & 67.88 & 63.66 \\
    \bottomrule
    \end{tabular}
    }
    \caption{Macro-F1 scores with different backbone models, on the HOVER dataset under open-book setting.}
    \label{tab:construction_model_comparison}
\end{table}


\paragraph{Generalizability Across Graph Construction Models}

Graph construction is a core component of GraphCheck, involving both latent entity detection and claim decomposition into factual triplets. To assess the generalizability of GraphCheck across different graph construction models, we explore (i) whether other LLMs can serve as effective alternatives to our default model (\texttt{\small gpt-4o}) through prompting alone, and (ii) whether relatively lightweight LLMs can achieve comparable results when fine-tuned.

We first examine the use of alternative LLMs through prompting alone. DP-GraphCheck maintains strong performance across different models, often matching or even exceeding that of \texttt{\small gpt-4o} (Table~\ref{tab:construction_model_comparison}). Although older or smaller models such as \texttt{\small gpt-3.5-turbo} and \texttt{\small Qwen2.5-72B-Instruct} exhibit relatively lower performance with GraphCheck alone, their performance improves substantially in DP-GraphCheck—reaching levels comparable to \texttt{\small gpt-4o}. This suggests that the strategy selector can effectively compensate for imperfect graph construction by leveraging Direct in cases where graph-based reasoning is deemed suboptimal.

Second, we investigate whether lightweight open-source models can serve as reliable graph constructors when fine-tuned. Specifically, we fine-tune \texttt{\small flan-t5-xl} (3B parameters) using pseudo-labels generated by \texttt{\small gpt-4o} (see Appx.\ref{appendix:graph_construction_ft} for training details). As shown in Table~\ref{tab:construction_model_comparison}, the fine-tuned \texttt{\small flan-t5-xl} achieves performance comparable to \texttt{\small gpt-4o}, particularly on 2-hop and 3-hop claims. This result demonstrates that properly fine-tuned smaller open-source models can serve as practical substitutes for large proprietary LLMs, avoiding privacy risks and high API costs.

Taken together, these findings show that the performance of GraphCheck and DP-GraphCheck with \texttt{\small gpt-4o} can be reproduced using other graph construction models—either by prompting alternative LLMs or fine-tuning smaller open-source models. Furthermore, the fact that DP-GraphCheck, even with smaller models, surpass the strongest baseline (DP-ProgramFC with \texttt{\small gpt-4o}) confirms that the strong performance of our approach primarily originates from its methodological advances.


\begin{table}[t!]
    \centering
    \renewcommand{\arraystretch}{1.3}
    \resizebox{\columnwidth}{!}{
    \begin{tabular}{llccc}
    \toprule
    \textbf{Method} & \textbf{Document-level Strategy} & \textbf{2-hop} & \textbf{3-hop} & \textbf{4-hop} \\
    \midrule
    \multirow{3}{*}{\textbf{Direct}} 
    & \textbf{concat}       & \textbf{72.56} & \textbf{61.70} & \textbf{59.57} \\
    & each         & 65.55 & 59.17 & 57.70 \\
    & concat+each & 67.90 & 59.92 & 58.40 \\
    \midrule
    \multirow{3}{*}{\textbf{GraphCheck}} 
    & concat       & 69.39 & 62.64 & 60.07 \\
    & each         & 74.79 & 66.19 & 62.36 \\
    & \textbf{concat+each} & \textbf{74.12} & \textbf{67.71} & \textbf{64.79} \\
    \bottomrule
    \end{tabular}
    }
    \caption{Macro-F1 scores of Direct and GraphCheck on HOVER across different document-level strategies under the open-book setting. Bold values denote the strategy 
    each method takes in the overall experiments.}
    \label{tab:verification_strategy}
\end{table}

\paragraph{Comparison of Document-level Strategies}

We compare three document-level strategies for claim verification with the top-$k$ retrieved documents: (1) \underline{\emph{concat}}, which verifies the claim against the concatenated documents; (2) \underline{\emph{each}}, which verifies the claim against each document individually and classifies it as \textsc{Supported} if any yields a \textsc{Supported} judgment; and (3) \underline{\emph{concat+each}}, which combines both approaches and classifies the claim as \textsc{Supported} if either the concatenated context or any individual document supports it. 

In DP-GraphCheck, Direct adopts the \emph{concat} strategy, while GraphCheck employs the \emph{concat+each}. As shown in Table~\ref{tab:verification_strategy}, these choices align with the nature of the input claims handled by each method. For Direct, the input is the original multi-hop claim, which integrates information across multiple documents. Verifying the claim against each document may fail to capture this broader context. Thus, evaluating over concatenated evidence (\emph{concat}) achieves the strongest performance within Direct. In contrast, GraphCheck verifies decomposed subclaims, each represented as an entity-relation triplet. Since these subclaims often correspond to atomic facts localized within individual documents, the \emph{each} strategy is well-suited. However, because some triplets still require cross-document context, the combined \emph{concat+each} strategy proves most effective for GraphCheck.

These results highlight the importance of aligning document-level strategies with the nature of the input—whether an original multi-hop claim or a decomposed triplet—for effective verification.


\paragraph{Experiment on FEVEROUS}

While FEVEROUS~\cite{aly-etal-2021-fact} does not fully align with our focus on higher-hop claims,
we include supplementary results in Appx.\ref{appendix:feverous} given its wide use in prior work. 
Recent baselines~\citep{pan-etal-2023-fact, wang2023explainable} limit retrieval to the introduction sections of Wikipedia articles, likely due to the length of full articles. However, since only 53.38\% of gold sentences appear in the introductions, this highly restricted setup is unrealistic. 
To address this issue, we expand the retrieval corpus to include the full article content.
Under this more realistic setting, GraphCheck again demonstrates superior performance over baseline methods.


\paragraph{Error Case Study}

To identify areas for improvement, we conduct an error analysis of GraphCheck. Specifically, we analyze 300 misclassified instances from the HOVER dataset, sampling 100 each from the 2-hop, 3-hop, and 4-hop subsets. Errors are categorized by pipeline stage to reveal the primary sources of failure. The results are summarized in Table~\ref{tab:error_analysis}.

We observe that errors from the graph construction stage are most prominent in low-hop claims, whereas errors in latent entity infilling increase significantly in higher-hop claims. Errors arising from fact triplet verification or labeling noise (i.e., incorrect ground-truth labels) remain relatively stable across hop levels. 

For graph construction errors, we further categorize them into three types (examples are provided in Appx.\ref{appendix:graph_error_examples}):
\begin{itemize}[itemsep=1pt, parsep=0pt, topsep=1pt]
    \item \textit{Hallucinated latent entities}: Non-latent entities are mistakenly marked as latent, potentially leading to faulty infilling.
    \item \textit{Missed latent entities}: Genuine latent entities are not detected. This case is relatively rare.
    \item \textit{Decomposition error}: Claim decomposition into sub-triplets results in semantic distortion or loss of key information.
\end{itemize}

Notably, in 2-hop claims, graph construction errors are dominant, with hallucinated latent entities being particularly frequent. This result indicates that the LLM tends to over-predict the presence of latent entities in simple claims, which often contain few or none.

In contrast, for higher-hop claims (i.e., 4-hop), latent entity infilling errors dominate, accounting for up to 54\% of errors. Since the hop count in HOVER corresponds to the number of gold documents, higher-hop claims inherently pose more challenging retrieval tasks. When essential documents are missing, the infilling step often fails to identify latent entities accurately, leading to incorrect claim verification.

This analysis highlights two directions for future improvements to GraphCheck: enhancing control of hallucinations in graph construction for simpler claims, and strengthening retrieval and infilling robustness for higher-hop claims.

\begin{table}[t!]
    \centering
    \renewcommand{\arraystretch}{1.2}
    \resizebox{\columnwidth}{!}{
    \begin{tabular}{lccc}
    \toprule
    \textbf{Error Type} & \textbf{2-hop} & \textbf{3-hop} & \textbf{4-hop} \\
    \midrule
    Graph Construction & 43\% & 27\% & 15\% \\
    \rowcolor{gray!10}
    \hspace{1em}- Hallucinated Latent Entities & 34\% & 14\% & 8\% \\
    \rowcolor{gray!10}
    \hspace{1em}- Missed Latent Entities & 1\% & 2\% & 1\% \\
    \rowcolor{gray!10}
    \hspace{1em}- Decomposition Error & 8\% & 11\% & 6\% \\
    Latent Entity Infilling & 11\% & 34\% & 54\% \\
    Fact Triplet Verification & 20\% & 21\% & 20\% \\
    Labeling Noise & 26\% & 18\% & 11\% \\
    \midrule
    \textbf{Total} & \textbf{100\%} & \textbf{100\%} & \textbf{100\%} \\
    \bottomrule
    \end{tabular}
    }
    \caption{Distribution of 300 error cases in GraphCheck on the HOVER dataset under open-book setting.}
    \label{tab:error_analysis}
\end{table}

\section{Conclusion}
We introduce GraphCheck, a novel framework for automated fact-checking based on entity-relationship graphs, specifically designed to handle multi-hop reasoning. By converting claims into structured entity-relationship graphs and exploring multiple identification paths, our approach enables comprehensive and systematic fact-checking. Additionally, we proposed DP-GraphCheck, which leverages a lightweight strategy selector to adaptively choose between simple direct prompting and more systematic graph-based fact-checking, thereby enhancing both efficiency and robustness.

Our experiments on the HOVER and EX-FEVER datasets demonstrate that DP-GraphCheck consistently achieves superior performance across multiple settings, while maintaining high computational efficiency. Furthermore, the strategy selector enhances the performance of other fact-checking methods, highlighting its broad applicability. Ablation studies validate the effectiveness of each component, including multipath exploration and document-level verification strategies. In addition, we confirm that GraphCheck generalizes across various graph construction models, including both alternative LLMs and fine-tuned smaller models. Overall, our findings highlight GraphCheck and DP-GraphCheck as strong and extensible frameworks for multi-hop fact-checking.

\section*{Limitations}
Despite its advantages, our framework has certain limitations. As shown in our error analysis, the construction of entity–relationship graphs can be error-prone, which may propagate to the verification stage. Common issues include misclassifying non-latent entities as latent and failing to decompose claims into a proper set of triplets.

Second, while our framework focuses on multi-hop fact-checking, it does not directly address multi-hop question answering, a widely studied task that also relies on multi-hop reasoning. Extending GraphCheck to this setting remains an open direction for future work.

Lastly, our framework currently operates solely over a textual knowledge base. Given that the verification process is grounded in a structured graph representation, future extensions could explore applying GraphCheck in settings where the evidence is sourced from structured knowledge bases such as knowledge graphs.

\bibliography{anthology, custom}

\begin{thebibliography}{28}
\providecommand{\natexlab}[1]{#1}

\bibitem[{Aly et~al.(2021)Aly, Guo, Schlichtkrull, Thorne, Vlachos,
  Christodoulopoulos, Cocarascu, and Mittal}]{aly-etal-2021-fact}
Rami Aly, Zhijiang Guo, Michael~Sejr Schlichtkrull, James Thorne, Andreas
  Vlachos, Christos Christodoulopoulos, Oana Cocarascu, and Arpit Mittal. 2021.
\newblock \href {https://doi.org/10.18653/v1/2021.fever-1.1} {The fact
  extraction and {VER}ification over unstructured and structured information
  ({FEVEROUS}) shared task}.
\newblock In \emph{Proceedings of the Fourth Workshop on Fact Extraction and
  VERification (FEVER)}, pages 1--13, Dominican Republic. Association for
  Computational Linguistics.

\bibitem[{Brown et~al.(2020)Brown, Mann, Ryder, Subbiah, Kaplan, Dhariwal,
  Neelakantan, Shyam, Sastry, Askell, Agarwal, Herbert-Voss, Krueger, Henighan,
  Child, Ramesh, Ziegler, Wu, Winter, Hesse, Chen, Sigler, Litwin, Gray, Chess,
  Clark, Berner, McCandlish, Radford, Sutskever, and
  Amodei}]{brown2020languagemodelsfewshotlearners}
Tom~B. Brown, Benjamin Mann, Nick Ryder, Melanie Subbiah, Jared Kaplan,
  Prafulla Dhariwal, Arvind Neelakantan, Pranav Shyam, Girish Sastry, Amanda
  Askell, Sandhini Agarwal, Ariel Herbert-Voss, Gretchen Krueger, Tom Henighan,
  Rewon Child, Aditya Ramesh, Daniel~M. Ziegler, Jeffrey Wu, Clemens Winter,
  and 12 others. 2020.
\newblock \href {https://arxiv.org/abs/2005.14165} {Language models are
  few-shot learners}.
\newblock \emph{Preprint}, arXiv:2005.14165.

\bibitem[{Chen and Shu(2023)}]{chen2023combatingmisinformationagellms}
Canyu Chen and Kai Shu. 2023.
\newblock \href {https://arxiv.org/abs/2311.05656} {Combating misinformation in
  the age of llms: Opportunities and challenges}.
\newblock \emph{Preprint}, arXiv:2311.05656.

\bibitem[{Chen et~al.(2021)Chen, Tworek, Jun, Yuan, de~Oliveira~Pinto, Kaplan,
  Edwards, Burda, Joseph, Brockman, Ray, Puri, Krueger, Petrov, Khlaaf, Sastry,
  Mishkin, Chan, Gray, Ryder, Pavlov, Power, Kaiser, Bavarian, Winter, Tillet,
  Such, Cummings, Plappert, Chantzis, Barnes, Herbert-Voss, Guss, Nichol,
  Paino, Tezak, Tang, Babuschkin, Balaji, Jain, Saunders, Hesse, Carr, Leike,
  Achiam, Misra, Morikawa, Radford, Knight, Brundage, Murati, Mayer, Welinder,
  McGrew, Amodei, McCandlish, Sutskever, and
  Zaremba}]{chen2021evaluatinglargelanguagemodels}
Mark Chen, Jerry Tworek, Heewoo Jun, Qiming Yuan, Henrique~Ponde
  de~Oliveira~Pinto, Jared Kaplan, Harri Edwards, Yuri Burda, Nicholas Joseph,
  Greg Brockman, Alex Ray, Raul Puri, Gretchen Krueger, Michael Petrov, Heidy
  Khlaaf, Girish Sastry, Pamela Mishkin, Brooke Chan, Scott Gray, and 39
  others. 2021.
\newblock \href {https://arxiv.org/abs/2107.03374} {Evaluating large language
  models trained on code}.
\newblock \emph{Preprint}, arXiv:2107.03374.

\bibitem[{Chung et~al.(2022)Chung, Hou, Longpre, Zoph, Tay, Fedus, Li, Wang,
  Dehghani, Brahma, Webson, Gu, Dai, Suzgun, Chen, Chowdhery, Castro-Ros,
  Pellat, Robinson, Valter, Narang, Mishra, Yu, Zhao, Huang, Dai, Yu, Petrov,
  Chi, Dean, Devlin, Roberts, Zhou, Le, and
  Wei}]{chung2022scalinginstructionfinetunedlanguagemodels}
Hyung~Won Chung, Le~Hou, Shayne Longpre, Barret Zoph, Yi~Tay, William Fedus,
  Yunxuan Li, Xuezhi Wang, Mostafa Dehghani, Siddhartha Brahma, Albert Webson,
  Shixiang~Shane Gu, Zhuyun Dai, Mirac Suzgun, Xinyun Chen, Aakanksha
  Chowdhery, Alex Castro-Ros, Marie Pellat, Kevin Robinson, and 16 others.
  2022.
\newblock \href {https://arxiv.org/abs/2210.11416} {Scaling
  instruction-finetuned language models}.
\newblock \emph{Preprint}, arXiv:2210.11416.

\bibitem[{Dmonte et~al.(2024)Dmonte, Oruche, Zampieri, Calyam, and
  Augenstein}]{dmonte2024claimverificationagelarge}
Alphaeus Dmonte, Roland Oruche, Marcos Zampieri, Prasad Calyam, and Isabelle
  Augenstein. 2024.
\newblock \href {https://arxiv.org/abs/2408.14317} {Claim verification in the
  age of large language models: A survey}.
\newblock \emph{Preprint}, arXiv:2408.14317.

\bibitem[{Gao et~al.(2024)Gao, Xiong, Gao, Jia, Pan, Bi, Dai, Sun, Wang, and
  Wang}]{gao2024retrievalaugmentedgenerationlargelanguage}
Yunfan Gao, Yun Xiong, Xinyu Gao, Kangxiang Jia, Jinliu Pan, Yuxi Bi, Yi~Dai,
  Jiawei Sun, Meng Wang, and Haofen Wang. 2024.
\newblock \href {https://arxiv.org/abs/2312.10997} {Retrieval-augmented
  generation for large language models: A survey}.
\newblock \emph{Preprint}, arXiv:2312.10997.

\bibitem[{Guo et~al.(2022)Guo, Schlichtkrull, and
  Vlachos}]{guo-etal-2022-survey}
Zhijiang Guo, Michael Schlichtkrull, and Andreas Vlachos. 2022.
\newblock \href {https://doi.org/10.1162/tacl_a_00454} {A survey on automated
  fact-checking}.
\newblock \emph{Transactions of the Association for Computational Linguistics},
  10:178--206.

\bibitem[{Jiang et~al.(2020)Jiang, Bordia, Zhong, Dognin, Singh, and
  Bansal}]{jiang-etal-2020-hover}
Yichen Jiang, Shikha Bordia, Zheng Zhong, Charles Dognin, Maneesh Singh, and
  Mohit Bansal. 2020.
\newblock \href {https://doi.org/10.18653/v1/2020.findings-emnlp.309}
  {{H}o{V}er: A dataset for many-hop fact extraction and claim verification}.
\newblock In \emph{Findings of the Association for Computational Linguistics:
  EMNLP 2020}, pages 3441--3460, Online. Association for Computational
  Linguistics.

\bibitem[{Khattab et~al.(2022)Khattab, Potts, and
  Zaharia}]{khattab2022baleenrobustmultihopreasoning}
Omar Khattab, Christopher Potts, and Matei Zaharia. 2022.
\newblock \href {https://arxiv.org/abs/2101.00436} {Baleen: Robust multi-hop
  reasoning at scale via condensed retrieval}.
\newblock \emph{Preprint}, arXiv:2101.00436.

\bibitem[{Lewis et~al.(2021)Lewis, Perez, Piktus, Petroni, Karpukhin, Goyal,
  Küttler, Lewis, tau Yih, Rocktäschel, Riedel, and
  Kiela}]{lewis2021retrievalaugmentedgenerationknowledgeintensivenlp}
Patrick Lewis, Ethan Perez, Aleksandra Piktus, Fabio Petroni, Vladimir
  Karpukhin, Naman Goyal, Heinrich Küttler, Mike Lewis, Wen tau Yih, Tim
  Rocktäschel, Sebastian Riedel, and Douwe Kiela. 2021.
\newblock \href {https://arxiv.org/abs/2005.11401} {Retrieval-augmented
  generation for knowledge-intensive nlp tasks}.
\newblock \emph{Preprint}, arXiv:2005.11401.

\bibitem[{Liu et~al.(2020)Liu, Xiong, Sun, and Liu}]{liu-etal-2020-fine}
Zhenghao Liu, Chenyan Xiong, Maosong Sun, and Zhiyuan Liu. 2020.
\newblock \href {https://doi.org/10.18653/v1/2020.acl-main.655} {Fine-grained
  fact verification with kernel graph attention network}.
\newblock In \emph{Proceedings of the 58th Annual Meeting of the Association
  for Computational Linguistics}, pages 7342--7351, Online. Association for
  Computational Linguistics.

\bibitem[{Ma et~al.(2024)Ma, Xu, Wei, Chen, Wang, Liu, Wu, and
  Wang}]{ma-etal-2024-ex}
Huanhuan Ma, Weizhi Xu, Yifan Wei, Liuji Chen, Liang Wang, Qiang Liu, Shu Wu,
  and Liang Wang. 2024.
\newblock \href {https://doi.org/10.18653/v1/2024.findings-acl.556}
  {{EX}-{FEVER}: A dataset for multi-hop explainable fact verification}.
\newblock In \emph{Findings of the Association for Computational Linguistics:
  ACL 2024}, pages 9340--9353, Bangkok, Thailand. Association for Computational
  Linguistics.

\bibitem[{Ostrowski et~al.(2021)Ostrowski, Arora, Atanasova, and
  Augenstein}]{ijcai2021p536}
Wojciech Ostrowski, Arnav Arora, Pepa Atanasova, and Isabelle Augenstein. 2021.
\newblock \href {https://doi.org/10.24963/ijcai.2021/536} {Multi-hop fact
  checking of political claims}.
\newblock In \emph{Proceedings of the Thirtieth International Joint Conference
  on Artificial Intelligence, {IJCAI-21}}, pages 3892--3898. International
  Joint Conferences on Artificial Intelligence Organization.
\newblock Main Track.

\bibitem[{Pan et~al.(2023{\natexlab{a}})Pan, Lu, Kan, and
  Nakov}]{pan-etal-2023-qacheck}
Liangming Pan, Xinyuan Lu, Min-Yen Kan, and Preslav Nakov. 2023{\natexlab{a}}.
\newblock \href {https://doi.org/10.18653/v1/2023.emnlp-demo.23} {{QAC}heck: A
  demonstration system for question-guided multi-hop fact-checking}.
\newblock In \emph{Proceedings of the 2023 Conference on Empirical Methods in
  Natural Language Processing: System Demonstrations}, pages 264--273,
  Singapore. Association for Computational Linguistics.

\bibitem[{Pan et~al.(2023{\natexlab{b}})Pan, Wu, Lu, Luu, Wang, Kan, and
  Nakov}]{pan-etal-2023-fact}
Liangming Pan, Xiaobao Wu, Xinyuan Lu, Anh~Tuan Luu, William~Yang Wang, Min-Yen
  Kan, and Preslav Nakov. 2023{\natexlab{b}}.
\newblock \href {https://doi.org/10.18653/v1/2023.acl-long.386} {Fact-checking
  complex claims with program-guided reasoning}.
\newblock In \emph{Proceedings of the 61st Annual Meeting of the Association
  for Computational Linguistics (Volume 1: Long Papers)}, pages 6981--7004,
  Toronto, Canada. Association for Computational Linguistics.

\bibitem[{Robertson and Zaragoza(2009)}]{10.1561/1500000019}
Stephen Robertson and Hugo Zaragoza. 2009.
\newblock \href {https://doi.org/10.1561/1500000019} {The probabilistic
  relevance framework: Bm25 and beyond}.
\newblock \emph{Found. Trends Inf. Retr.}, 3(4):333–389.

\bibitem[{Shao et~al.(2023)Shao, Gong, Shen, Huang, Duan, and
  Chen}]{shao-etal-2023-enhancing}
Zhihong Shao, Yeyun Gong, Yelong Shen, Minlie Huang, Nan Duan, and Weizhu Chen.
  2023.
\newblock \href {https://doi.org/10.18653/v1/2023.findings-emnlp.620}
  {Enhancing retrieval-augmented large language models with iterative
  retrieval-generation synergy}.
\newblock In \emph{Findings of the Association for Computational Linguistics:
  EMNLP 2023}, pages 9248--9274, Singapore. Association for Computational
  Linguistics.

\bibitem[{Thorne et~al.(2018)Thorne, Vlachos, Christodoulopoulos, and
  Mittal}]{thorne-etal-2018-fever}
James Thorne, Andreas Vlachos, Christos Christodoulopoulos, and Arpit Mittal.
  2018.
\newblock \href {https://doi.org/10.18653/v1/N18-1074} {{FEVER}: a large-scale
  dataset for fact extraction and {VER}ification}.
\newblock In \emph{Proceedings of the 2018 Conference of the North {A}merican
  Chapter of the Association for Computational Linguistics: Human Language
  Technologies, Volume 1 (Long Papers)}, pages 809--819, New Orleans,
  Louisiana. Association for Computational Linguistics.

\bibitem[{Vlachos and Riedel(2014)}]{vlachos-riedel-2014-fact}
Andreas Vlachos and Sebastian Riedel. 2014.
\newblock \href {https://doi.org/10.3115/v1/W14-2508} {Fact checking: Task
  definition and dataset construction}.
\newblock In \emph{Proceedings of the {ACL} 2014 Workshop on Language
  Technologies and Computational Social Science}, pages 18--22, Baltimore, MD,
  USA. Association for Computational Linguistics.

\bibitem[{Wang and Shu(2023)}]{wang2023explainable}
Haoran Wang and Kai Shu. 2023.
\newblock Explainable claim verification via knowledge-grounded reasoning with
  large language models.
\newblock \emph{arXiv preprint arXiv:2310.05253}.

\bibitem[{Wang(2017)}]{wang-2017-liar}
William~Yang Wang. 2017.
\newblock \href {https://doi.org/10.18653/v1/P17-2067}
  {{\textquotedblleft}liar, liar pants on fire{\textquotedblright}: A new
  benchmark dataset for fake news detection}.
\newblock In \emph{Proceedings of the 55th Annual Meeting of the Association
  for Computational Linguistics (Volume 2: Short Papers)}, pages 422--426,
  Vancouver, Canada. Association for Computational Linguistics.

\bibitem[{Yang et~al.(2018)Yang, Qi, Zhang, Bengio, Cohen, Salakhutdinov, and
  Manning}]{yang-etal-2018-hotpotqa}
Zhilin Yang, Peng Qi, Saizheng Zhang, Yoshua Bengio, William Cohen, Ruslan
  Salakhutdinov, and Christopher~D. Manning. 2018.
\newblock \href {https://doi.org/10.18653/v1/D18-1259} {{H}otpot{QA}: A dataset
  for diverse, explainable multi-hop question answering}.
\newblock In \emph{Proceedings of the 2018 Conference on Empirical Methods in
  Natural Language Processing}, pages 2369--2380, Brussels, Belgium.
  Association for Computational Linguistics.

\bibitem[{Yuan and Vlachos(2024)}]{yuan-vlachos-2024-zero}
Moy Yuan and Andreas Vlachos. 2024.
\newblock \href {https://doi.org/10.18653/v1/2024.kallm-1.11} {Zero-shot
  fact-checking with semantic triples and knowledge graphs}.
\newblock In \emph{Proceedings of the 1st Workshop on Knowledge Graphs and
  Large Language Models (KaLLM 2024)}, pages 105--115, Bangkok, Thailand.
  Association for Computational Linguistics.

\bibitem[{Zhang and Gao(2023)}]{zhang-gao-2023-towards}
Xuan Zhang and Wei Gao. 2023.
\newblock \href {https://doi.org/10.18653/v1/2023.ijcnlp-main.64} {Towards
  {LLM}-based fact verification on news claims with a hierarchical step-by-step
  prompting method}.
\newblock In \emph{Proceedings of the 13th International Joint Conference on
  Natural Language Processing and the 3rd Conference of the Asia-Pacific
  Chapter of the Association for Computational Linguistics (Volume 1: Long
  Papers)}, pages 996--1011, Nusa Dua, Bali. Association for Computational
  Linguistics.

\bibitem[{Zhong et~al.(2020)Zhong, Xu, Tang, Xu, Duan, Zhou, Wang, and
  Yin}]{zhong-etal-2020-reasoning}
Wanjun Zhong, Jingjing Xu, Duyu Tang, Zenan Xu, Nan Duan, Ming Zhou, Jiahai
  Wang, and Jian Yin. 2020.
\newblock \href {https://doi.org/10.18653/v1/2020.acl-main.549} {Reasoning over
  semantic-level graph for fact checking}.
\newblock In \emph{Proceedings of the 58th Annual Meeting of the Association
  for Computational Linguistics}, pages 6170--6180, Online. Association for
  Computational Linguistics.

\bibitem[{Zhou et~al.(2019)Zhou, Han, Yang, Liu, Wang, Li, and
  Sun}]{zhou-etal-2019-gear}
Jie Zhou, Xu~Han, Cheng Yang, Zhiyuan Liu, Lifeng Wang, Changcheng Li, and
  Maosong Sun. 2019.
\newblock \href {https://doi.org/10.18653/v1/P19-1085} {{GEAR}: Graph-based
  evidence aggregating and reasoning for fact verification}.
\newblock In \emph{Proceedings of the 57th Annual Meeting of the Association
  for Computational Linguistics}, pages 892--901, Florence, Italy. Association
  for Computational Linguistics.

\bibitem[{Zhu et~al.(2019)Zhu, Hu, and Xing}]{zhu2019textinfilling}
Wanrong Zhu, Zhiting Hu, and Eric Xing. 2019.
\newblock \href {https://arxiv.org/abs/1901.00158} {Text infilling}.
\newblock \emph{Preprint}, arXiv:1901.00158.

\end{thebibliography}

\appendix

\section{Comparison with a Fine-tuned Fact-Checking Model}
\label{appendix:ft_model}

While our main experiments use methods that leverage the in-context learning capabilities of LLMs without task-specific training as baselines, we additionally compare GraphCheck with fine-tuned models for a more comprehensive evaluation. Specifically, we employ the BERT model fine-tuned for the fact-checking task by \citet{jiang-etal-2020-hover}.

\paragraph{Gold Setting}

Following the experimental setup of \citet{jiang-etal-2020-hover}, we compare our approach with fine-tuned BERT in the \textit{gold} setting, in which verification is based exclusively on gold evidence. This setting differs from the \textit{open-book + gold} setting used in our main experiments, where gold evidence is combined with retrieved documents.

Table~\ref{tab:bert_comparison_gold} reports accuracy under the gold setting. Although BERT generally achieves higher accuracy, DP-GraphCheck surpasses it on 2-hop claims, relying only on 10 few-shot examples from HOVER, whereas BERT requires supervised training on the full training set.

\begin{table}[h!]
    \centering
    \renewcommand{\arraystretch}{1.2}
    \small
    \begin{tabular}{lccc}
    \toprule
    \textbf{Method} & \textbf{2-hop} & \textbf{3-hop} & \textbf{4-hop} \\
    \midrule
    BERT* & 79.8 & \textbf{83.5} & \textbf{78.6} \\
    GraphCheck & 78.8 & 73.2 & 72.0 \\
    DP-GraphCheck & \textbf{81.7} & 73.8 & 74.1 \\
    \bottomrule
    \end{tabular}
    \caption{Accuracy on the HOVER dataset under gold setting. The best result in each column is highlighted in \textbf{bold}. \textsuperscript{*}BERT results reported by \citet{jiang-etal-2020-hover}.}
    \label{tab:bert_comparison_gold}
\end{table}

\paragraph{Open-book + Gold Setting}

In the more challenging open-book + gold setting where gold documents are merged with retrieved documents, the performance of fine-tuned BERT degrades significantly due to noise introduced by retrieved documents (Table~\ref{tab:bert_comparison_noisy}). In contrast, DP-GraphCheck maintains robust performance, demonstrating superior noise tolerance.

\begin{table}[h!]
    \centering
    \renewcommand{\arraystretch}{1.2}
    \small
    \begin{tabular}{lcccc}
    \toprule
    \textbf{Method} & \textbf{2-hop} & \textbf{3-hop} & \textbf{4-hop} \\
    \midrule
    BERT & 72.5 & 69.0 & 62.7 \\
    GraphCheck & \textbf{78.7} & \textbf{73.8} & 73.0 \\
    DP-GraphCheck & 77.1 & 73.7 & \textbf{73.6} \\
    \bottomrule
    \end{tabular}
    \caption{Accuracy on the HOVER dataset under open-book + gold setting. The best result in each column is highlighted in \textbf{bold}. Note that the table reports accuracy for consistency with prior work~\citep{jiang-etal-2020-hover}, whereas our main results use macro-F1 scores.}
    \label{tab:bert_comparison_noisy}
\end{table}

\paragraph{Cross-domain Generalization}

Fine-tuned models often face challenges in domain adaptation. As shown in Table~\ref{tab:cross_domain}, BERT fine-tuned on HOVER experiences a substantial performance drop when evaluated on EX-FEVER. In contrast, DP-GraphCheck demonstrates strong cross-domain generalization using the same few-shot prompts derived solely from HOVER, without any exposure to EX-FEVER samples during inference.

\begin{table}[h!]
    \centering
    \renewcommand{\arraystretch}{1.2}
    \small
    \begin{tabular}{lccc}
    \toprule
    \textbf{Method} & \textbf{2-hop} & \textbf{3-hop} \\
    \midrule
    BERT & 60.8 & 54.3 \\
    GraphCheck & 83.7 & 80.2 \\
    DP-GraphCheck & \textbf{85.7} & \textbf{82.9} \\
    \bottomrule
    \end{tabular}
    \caption{Accuracy on the EX-FEVER dataset using models trained/prompted with the HOVER dataset (open-book + gold setting). The best result in each column is highlighted in \textbf{bold}. Note that the table reports accuracy for consistency with prior work~\citep{jiang-etal-2020-hover}, whereas our main results use macro-F1 scores.}
    \label{tab:cross_domain}
\end{table}

Overall, these results highlight the competitive performance and robust generalization ability of GraphCheck, despite requiring no task-specific fine-tuning. While fine-tuned models can perform well in controlled settings, our approach provides a more practical solution for real-world applications where training data may be limited, domains may shift, and retrieval is noisy.

\section{Evaluation on FEVEROUS}
\label{appendix:feverous}

\begin{table}[h!]
    \centering
    \renewcommand{\arraystretch}{1.2}
    \small
    \resizebox{\columnwidth}{!}{
    \begin{tabular}{lcc}
    \toprule
    \textbf{Method} & \textbf{FEVEROUS-Intro} & \textbf{FEVEROUS-Alpha} \\
    \midrule
    Direct & \textbf{69.48} & 85.28 \\
    ProgramFC ($N=5$) & 66.49 & 84.17 \\
    DP-ProgramFC ($N=5$) & 66.11 & 84.20 \\
    FOLK & 61.31 & 76.78 \\
    DP-FOLK & 66.39 & 84.71 \\
    GraphCheck ($\bar{P}=5$) & 63.06 & 85.37 \\
    DP-GraphCheck ($\bar{P}=5$) & 66.34 & \textbf{86.87} \\
    \bottomrule
    \end{tabular}
    }
    \caption{Macro-F1 scores on the FEVEROUS dataset under open-book setting.}
\label{tab:feverous_results}
\end{table}

Although FEVEROUS~\cite{aly-etal-2021-fact} is widely used in related work, we do not include it in our main results because FEVEROUS is primarily designed for multi-hop reasoning that combines table and text evidence, whereas our focus is on higher-hop claims with purely textual evidence. Indeed, when limited to text-only, FEVEROUS predominantly contains relatively low-hop claims. 

However, to ensure a comprehensive evaluation, we conduct supplementary experiments. Since our approach focuses on textual verification, we extract a subset containing only claims with textual evidence, following ProgramFC~\cite{pan-etal-2023-fact}. As test labels are not publicly released, we use the development set (2,962 instances) for evaluation and randomly sample 2,000 instances from the training set for validation. We conduct validation experiments to determine the most suitable document-level strategy (concat, each, concat+each) and choose the concat for evaluation. 

Regarding the retrieval corpus, we utilize the preprocessed Wikipedia dump from December 2020. Recent baselines~\citep{pan-etal-2023-fact, wang2023explainable} typically limit retrieval to the introduction section of each Wikipedia article, likely due to the excessive length of full articles. We refer to this setup as \textit{FEVEROUS-Intro}. However, this design choice restricts evidence coverage—only 53.38\% of gold evidence sentences are found within introductions. To mitigate this issue, we expand the corpus to include the full article content, segmented into three-sentence chunks to improve both coverage and retrieval granularity. We refer to this enhanced setup as \textit{FEVEROUS-Alpha}.

The results are reported in Table~\ref{tab:feverous_results}. In the FEVEROUS-Intro setting, Direct achieves the highest score, possibly because it compensates for limited evidence coverage using its parametric knowledge. However, we believe this setup—where the retrieval corpus itself is highly restricted—is unrealistic. In the more realistic FEVEROUS-Alpha setting, DP-GraphCheck achieves the best performance, demonstrating its effectiveness on the FEVEROUS dataset.

\section{Analysis of Hop Count and Latent Entities}
\label{appendix:num_hops_analysis}

On the HOVER dataset, we observe a strong correlation between hop count and the number of latent entities in the constructed graphs (Table~\ref{tab:latent_entity_distribution}). As graphs with more latent entities yield more possible identification orders, GraphCheck allocates fewer paths to lower-hop claims and more paths to higher-hop claims. This indicates that GraphCheck adaptively adjusts path exploration by claim complexity, unlike prior methods that assign a fixed number of paths to all claims.

\begin{table}[h!]
    \centering
    \renewcommand{\arraystretch}{1.2}
    \small
    \begin{tabular}{lccc}
    \toprule
    \textbf{\# of latent entities} & \textbf{2-hop} & \textbf{3-hop} & \textbf{4-hop} \\
    \midrule
    \textbf{0} & 4.3\% & 0.1\% & 0.0\% \\
    \textbf{1} & 63.0\% & 44.0\% & 3.3\% \\
    \textbf{2} & 25.1\% & 37.8\% & 59.6\% \\
    \textbf{3+} & 7.6\% & 18.1\% & 37.2\% \\
    \midrule
    \textbf{Total} & 100.0\% & 100.0\% & 100.0\% \\
    \bottomrule
    \end{tabular}
    \caption{Distribution of the number of latent entities across hop levels in the HOVER dataset.}
    \label{tab:latent_entity_distribution}
\end{table}

\section{Fine-tuning for Graph Construction}
\label{appendix:graph_construction_ft}

To reduce reliance on large proprietary models, we train a smaller open-source model, \texttt{\small flan-t5-xl}, for the graph construction task. 

We use graphs initially generated by \texttt{\small gpt-4o} as pseudo-labels, but retain only those graphs that enable GraphCheck to predict the correct claim veracity. These filtered graphs then serve as supervision for training. 

The pseudo-labeled training data consists of claims from the HOVER, EX-FEVER, and FEVER training splits, with 3,000 HOVER claims reserved for hyperparameter tuning and dataset ratio adjustments. We incorporate the single-hop FEVER dataset, which rarely contains latent entities, to mitigate hallucinated latent entity errors that frequently occur in GPT-based construction.

We also decompose graph construction into two subtasks: detecting latent entities within claims and extracting fact triplets. We train a separate model for each subtask, allowing more focused learning for latent entity detection and triplet extraction. Fine-tuning is performed with a learning rate of $5\times10^{-4}$ for 3 epochs with early stopping.

\section{DP-GraphCheck Algorithm}
\label{appendix:algorithm}

\begin{flushright}
\begin{minipage}{\columnwidth}
\raggedright
\algrenewcommand{\alglinenumber}[1]{\fontsize{7}{10}\selectfont #1:\quad}
\captionsetup[algorithm]{labelfont={footnotesize,bf}, font=footnotesize}
\vspace{-10pt}
\begin{algorithm}[H]
\caption{DP-GraphCheck}
\label{Algorithm 1}
\fontsize{9}{12}\selectfont
    \begin{algorithmic}
        \State \textbf{Input:} Claim $C$
        \State \textbf{Modules:} Strategy Selector $\mathcal{S}$, Verifier $\mathcal{V}$, Retriever $\mathcal{R}$

        \State Retrieve top-$k$ documents $\{d_1, \dots, d_k\} \gets \mathcal{R}(C)$
        \State Concatenate docs: $d_{\text{concat}} \gets d_1 \oplus \cdots \oplus d_k$
        \State Determine strategy $\sigma \gets \mathcal{S}(C, d_{\text{concat}})$
        
        \If{$\sigma = \textsc{Direct}$}
            \State Use verifier $\mathcal{V}(C, d_{\text{concat}})$ to classify $C$
            \State \textbf{return} $\mathcal{V}(C, d_{\text{concat}})$
        \EndIf

        \State \textbf{\# GraphCheck}
        \State Construct graph $\mathcal{G}_C = \{t_1, t_2, \dots, t_n\}$
        
        \For{\textbf{each} path $\pi_p \in \{\pi_1, \dots, \pi_{\min(P, \bar{P})}\}$}
            \State Infill graph $\mathcal{G}_{C,\pi_p}' = \{t_1', t_2', \dots, t_n'\}$
            \State $path\_supported \gets$ \textbf{True}
            
            \For{\textbf{each} infilled triplet $t_i' \in \mathcal{G}_{C,\pi_p}'$}
                \State Retrieve top-$k$ documents $\{d_1, \dots, d_k\} \gets \mathcal{R}(t_i')$
                \State Concatenate docs: $d_{\text{concat}} \gets d_1 \oplus \cdots \oplus d_k$
		      \State $triplet\_supported \gets$ \textbf{False}
                \For{\textbf{each} $d_j \in \{d_{\text{concat}}, d_1, \dots, d_k\}$}
                    \If{$\mathcal{V}(t_i', d_j) = \textsc{Supported}$}
                        \State $triplet\_supported \gets$ \textbf{True}
                        \State \textbf{break}
                    \EndIf
                \EndFor
                \If{\textbf{not} $triplet\_supported$}
                    \State $path\_supported \gets$ \textbf{False}
                    \State \textbf{break}
                \EndIf
            \EndFor
            
            \If{$path\_supported$}
                \State \textbf{return} \textsc{Supported}
            \EndIf
        \EndFor
        
        \State \textbf{return} \textsc{Not Supported}
    \end{algorithmic}
\end{algorithm}
\end{minipage}
\end{flushright}

\newpage
\section{An Example from the EX-FEVER Dataset}
\label{appendix:ex-fever_example}

\begin{figure}[h!]
    \centering
    \includegraphics[width=\columnwidth]{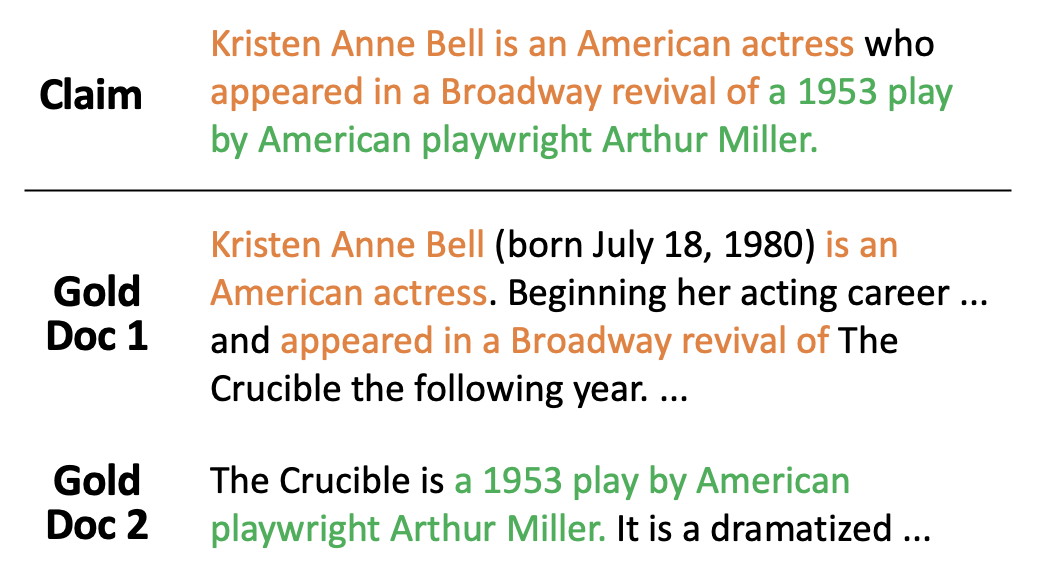}
    \caption{An example from the EX-FEVER dataset.}
    \label{fig:ex-fever_example}
\end{figure}

EX-FEVER is highly extractive compared to the HOVER dataset. As illustrated in the example shown in Figure \ref{fig:ex-fever_example}, claims in EX-FEVER closely match phrases in the gold documents with minimal rephrasing or abstraction.

\section{Dataset Statistics and Model Specifications}
\label{appendix:details}

\subsection{Dataset Statistics} 

\textbf{HOVER}~\cite{jiang-etal-2020-hover} consists of 18,171 training, 4,000 development, and 4,000 test claims, requiring 2 to 4 hops of reasoning. Labels are \textsc{Supported} or \textsc{Not Supported}. Since test labels are not publicly available, we use the development set for evaluation. 

\textbf{EX-FEVER}~\cite{ma-etal-2024-ex} contains 43,107 training, 12,059 development, and 6,099 test claims, with 2 to 3 hop reasoning. Labels are \textsc{Supports}, \textsc{Refutes}, and \textsc{Not Enough Info (NEI)}. Based on discussions with the dataset creators, we confirm that the NEI label does not always indicate a true absence of evidence in the retrieval corpus, but only in the annotated subset. In open-domain retrieval settings, NEI-labeled claims can often be verified with evidence, making NEI label less reliable for evaluation. We therefore exclude NEI samples from our experiments. 

\textbf{FEVEROUS}~\cite{aly-etal-2021-fact} consists of 71,291 training, 7,890 development, and 7,845 test claims. Labels are \textsc{Supported}, \textsc{Refuted}, and \textsc{NEI}. FEVEROUS involves evidence from both unstructured text and structured tables. Since we focus on textual fact-checking, we adopt the subset of \citet{pan-etal-2023-fact}, which selects only claims that require sentence-only evidence from the development set, yielding 2,962 claims for evaluation. 

\textbf{FEVER}~\cite{thorne-etal-2018-fever} is a large-scale single-hop fact-checking dataset with 145,449 training, 9,999 development, and 9,999 test claims. Labels are \textsc{Supported}, \textsc{Refuted}, and \textsc{NEI}. The dataset consists mostly of simple claims that require single-hop reasoning, such as ``The Beatles was a rock band''.

\subsection{Model Specifications}

We experiment with a range of LLMs of different scales and access types. 

For open-source models, we use \texttt{\small flan-t5-xl} (3B parameters) and \texttt{\small Qwen2.5-72B-Instruct} (72B parameters), both released on Hugging Face.

Among API-based proprietary models, we use \texttt{\small gpt-4o-2024-08-06} and \texttt{\small gpt-3.5-turbo} from OpenAI, whose parameter counts and architectural details have not been publicly disclosed. We further include \texttt{\small claude-3.7-sonnet-20250219} from Anthropic, also without public parameter specifications.

\subsection{Licenses and Terms of Use}

HOVER is released under the CC BY-SA 4.0 license, while EX-FEVER is provided for research purposes. FEVER is distributed under the CC BY-SA 3.0 and GPL-3.0 licenses, and FEVEROUS is released under the CC BY-SA 4.0 license. \texttt{\small flan-t5-xl} and \texttt{\small Qwen2.5-72B-Instruct} are available on Hugging Face under the Apache 2.0 and Qwen license, respectively. API-based proprietary models (\texttt{\small gpt-4o-2024-08-06}, \texttt{\small gpt-3.5-turbo}, and \texttt{\small claude-3.7-sonnet-20250219}) are accessed in accordance with their respective terms of use. All artifacts are used strictly for research purposes and in a manner consistent with their intended use and license conditions.

\onecolumn
\section{Prompts}
\label{appendix:prompts}

\subsection{Graph Construction}
\label{appendix:prompt_graph_construction}

The prompt in Table~\ref{tab:prompt_graph_construction} is used in GraphCheck to convert a textual claim into a structured entity-relationship graph. It consists of an instruction segment and 10 illustrative examples manually created using claims from the HOVER training set.

\begin{small}
\begin{longtable}{|p{0.95\textwidth}|}
    \hline
    \rule{0pt}{1.5em}
    We are conducting fact-checking on multi-hop claims. To facilitate this process, we need to decompose each claim into triples for more granular and accurate fact-checking. Please follow the guidelines below when decomposing claims into triples:
    
    \textit{\# Latent Entities:}
    \begin{itemize}[nosep, left=0.5em]
        \item (Identification) Firstly, identify any latent entities (i.e., implicit references not directly mentioned in the claim) that need to be clarified for accurate fact-checking.
        \item (Definition) Define these identified latent entities in triple format, using placeholders like (ENT1), (ENT2), etc.
    \end{itemize}
    
    \textit{\# Triples:}
    \begin{itemize}[nosep, left=0.5em]
        \item (Basic Information Unit) Decompose the claim into triples, ensuring you reach the most fundamental verifiable information while preserving the original meaning. Be careful not to lose important information during decomposition.
        \item (Triple Structure) Each triple should follow this format: `subject [SEP] relation [SEP] object'. Both the subject and object should be noun phrases, while the relation should be a verb or verb phrase, forming a complete sentence.
        \item (Prepositional Phrases) In exceptional cases where a prepositional phrase modifies the entire triple (rather than just the subject or object) and splitting it into another triple would alter the meaning of the claim, do not divide it. Instead, append it to the end of the triple: `subject [SEP] relation [SEP] object [PREP] preposition phrase'.
        \item (Pronoun Resolution) Replace any pronouns with the corresponding entities to ensure that each triple is self-contained and independent of external context.
        \item (Entity Consistency) Use the exact same string to represent entities (i.e., the `subject' or `object') whenever they refer to the same entity across different triples.
    \end{itemize}
    
    \vspace{1em}
    \textit{\# Claim:} \\
    The fairy Queen Mab orginated with William Shakespeare. \\
    \textit{\# Latent Entities:} \\
    \textit{\# Triples:} \\
    The fairy Queen Mab [SEP] originated with [SEP] William Shakespeare\\\\
    
    \textit{\# Claim:} \\
    Giacomo Benvenuti and Claudio Monteverdi share the profession of Italian composer.\\
    \textit{\# Latent Entities:} \\
    \textit{\# Triples:} \\
    Giacomo Benvenuti [SEP] is [SEP] Italian composer\\
    Claudio Monteverdi [SEP] is [SEP] Italian composer\\\\
    
    \textit{\# Claim:} \\
    Ross Pople worked with the English composer Michael Tippett, who is known for his opera ``The Midsummer Marriage''.\\
    \textit{\# Latent Entities:} \\
    \textit{\# Triples:} \\
    Ross Pople [SEP] worked with [SEP] the English composer Michael Tippett\\
    The English composer Michael Tippett [SEP] is known for [SEP] the opera ``The Midsummer Marriage''\\\\
    
    \textit{\# Claim:} \\
    Mark Geragos was involved in the scandal that took place in the 1990s.\\
    \textit{\# Latent Entities:} \\
    (ENT1) [SEP] is [SEP] a scandal\\
    \textit{\# Triples:} \\
    Mark Geragos [SEP] was involved in [SEP] (ENT1)\\
    (ENT1) [SEP] took place in [SEP] the 1990s\\\\
    
    \textit{\# Claim:} \\
    Where is the airline company that operated United Express Flight 3411 on April 9, 2017 on behalf of United Express is headquartered in Indianapolis, Indiana.\\
    \textit{\# Latent Entities:} \\
    (ENT1) [SEP] is [SEP] an airline company\\
    \textit{\# Triples:} \\
    (ENT1) [SEP] operated [SEP] United Express Flight 3411 [PREP] on April 9, 2017 on behalf of United Express\\
    (ENT1) [SEP] is headquartered in [SEP] Indianapolis, Indiana\\\\
    
    \textit{\# Claim:} \\
    The Skatoony has reruns on Teletoon in Canada and was shown between midnight and 6:00 on the network that launched 24 April 2006, the same day as rival Nick Jr. Too.\\
    \textit{\# Latent Entities:} \\
    (ENT1) [SEP] is [SEP] a network\\
    \textit{\# Triples:} \\
    Skatoony [SEP] has reruns on [SEP] Teletoon\\
    Teletoon [SEP] is located in [SEP] Canada\\
    Skatoony [SEP] was shown on [SEP] (ENT1) [PREP] between midnight and 6:00\\
    (ENT1) [SEP] launched on [SEP] 24 April 2006\\
    Nick Jr. Too [SEP] launched on [SEP] 24 April 2006\\\\
    
    \textit{\# Claim:} \\
    Danny Shirley is older than Kevin Parker.\\
    \textit{\# Latent Entities:} \\
    (ENT1) [SEP] is [SEP] a date\\
    (ENT2) [SEP] is [SEP] a date\\
    \textit{\# Triples:} \\
    Danny Shirley [SEP] was born on [SEP] (ENT1)\\
    Kevin Parker [SEP] was born on [SEP] (ENT2)\\
    (ENT1) [SEP] is before [SEP] (ENT2)\\\\
    
    \textit{\# Claim:} \\
    The founder of this Canadian owned, American manufacturer of business jets for civilian and military did not develop the 8-track portable tape system.\\
    \textit{\# Latent Entities:} \\
    (ENT1) [SEP] is [SEP] an individual\\
    (ENT2) [SEP] is [SEP] an American manufacturer\\
    \textit{\# Triples:} \\
    (ENT1) [SEP] founded [SEP] (ENT2)\\
    (ENT2) [SEP] is owned by [SEP] Canadian\\
    (ENT2) [SEP] made [SEP] business jets for civilian and military\\
    (ENT1) [SEP] did not develop [SEP] 8-track portable tape system\\\\
    
    \textit{\# Claim:} \\
    The Dutch man who along with Dennis Bergkamp was acquired in the 1993-94 Inter Milan season, manages Cruyff Football together with the footballer who is also currently manager of Tel Aviv team.\\
    \textit{\# Latent Entities:} \\
    (ENT1) [SEP] is [SEP] a Dutch man\\
    (ENT2) [SEP] is [SEP] a footballer\\
    \textit{\# Triples:} \\
    (ENT1) [SEP] was acquired in [SEP] the 1993-94 Inter Milan season [PREP] along with Dennis Bergkamp\\
    (ENT1) [SEP] manages [SEP] Cruyff Football [PREP] together with (ENT2)\\
    (ENT2) [SEP] currently manages [SEP] Tel Aviv team\\\\
    
    \textit{\# Claim:} \\
    An actor starred in the 2007 film based on a former FBI agent. That agent was Robert Philip Hanssen. The actor starred in the 2005 Capitol film Chaos.\\
    \textit{\# Latent Entities:} \\
    (ENT1) [SEP] is [SEP] an actor\\
    (ENT2) [SEP] is [SEP] a 2007 film\\
    \textit{\# Triples:} \\
    (ENT1) [SEP] starred in [SEP] (ENT2)\\
    (ENT2) [SEP] is based on [SEP] Robert Philip Hanssen\\
    Robert Philip Hanssen [SEP] is [SEP] a former FBI agent\\
    (ENT1) [SEP] starred in [SEP] the 2005 Capitol film Chaos\\\\
    
    \textit{\# Claim:} \\
    \texttt{\{claim\}} \\\\
    
    \hline
    \caption{Prompt used for graph construction.}
    \label{tab:prompt_graph_construction}
\end{longtable}
\end{small}


\newpage
\subsection{Latent Entity Infilling}
\label{appendix:prompt_infilling}

The prompt in Table~\ref{tab:prompt_infilling} is used in the latent entity infilling step of GraphCheck to identify the target latent entity.

\begin{table}[h]
    \centering
    \renewcommand{\arraystretch}{1.2}
    \small
    \begin{tabular}{|p{0.95\textwidth}|}
    \hline
    \vspace{0.01em}
    \texttt{\{context\}}\\
    Based on the above information, fill in the blank with the correct entity: \texttt{\{infilling\_query\}}\\
    \textit{Answer:} \\[0.5em]
    \hline
    \end{tabular}
    \caption{Prompt used for latent entity infilling.}
    \label{tab:prompt_infilling}
\end{table}

The \texttt{\small \{context\}} refers to the concatenation of top-$k$ documents retrieved using a custom retrieval query. This retrieval query is constructed by concatenating all triplets in the \textit{\# Triples} section that include the target latent entity to be infilled and exclude any other unidentified latent entities. In this retrieval query, the placeholder for the target latent entity is replaced with the corresponding textual reference, as specified in the \textit{\# Latent Entities} section. All \texttt{\small [SEP]} tokens are removed.

The \texttt{\small \{infilling\_query\}} is constructed by concatenating all relevant triplets from both the \textit{\# Latent Entities} and \textit{\# Triples} sections that include the target latent entity and exclude any other unidentified latent entities. In this case, the placeholder for the target latent entity is replaced with the special token \texttt{\small <extra\_id\_0>} to indicate the blank to be infilled. As above, all \texttt{\small [SEP]} tokens are removed.

See Table~\ref{tab:example_infilling_queries} for a detailed example of how the retrieval and infilling queries are constructed.

\begin{table}[h]
    \centering
    \renewcommand{\arraystretch}{1.2}
    \small
    \begin{tabular}{p{0.95\textwidth}}
    \hline
    \vspace{0.01em}
    Consider the following graph:
    
    \vspace{0.5em}
    \textit{\# Latent Entities:} \\
    (ENT1) [SEP] is [SEP] a musician \\
    (ENT2) [SEP] is [SEP] a band \\
    \textit{\# Triples:} \\
    (ENT1) [SEP] is part of [SEP] Tall Birds \\
    (ENT1) [SEP] is a percussionist for [SEP] (ENT2) \\
    (ENT2) [SEP] formed in [SEP] Issaquah, Washington \\
    
    \vspace{0.5em}
    In the case where (ENT1) is the first target latent entity to be infilled, the corresponding retrieval and infilling queries are:
    
    \begin{itemize}[nosep, left=0.5em]
        \item \textit{Retrieval Query:} a musician is part of Tall Birds.
        \item \textit{Infilling Query:} <extra\_id\_0> is part of Tall Birds. <extra\_id\_0> is a musician.
    \end{itemize}
    
    \vspace{1em}
    After (ENT1) is identified as \textit{``Randall Nieman''}, (ENT2) becomes the next target latent entity to be infilled. The corresponding retrieval and infilling queries are:
    
    \begin{itemize}[nosep, left=0.5em]
        \item \textit{Retrieval Query:} Randall Nieman is a percussionist for a band. a band formed in Issaquah, Washington.
        \item \textit{Infilling Query:} Randall Nieman is a percussionist for <extra\_id\_0>. <extra\_id\_0> formed in Issaquah, Washington. <extra\_id\_0> is a band.
    \end{itemize}
    \\\\\hline
    \end{tabular}
    \caption{Example of retrieval and infilling queries for latent entity infilling.}
    \label{tab:example_infilling_queries}
\end{table}


\subsection{Direct (Fact Triplet Verification)}
\label{appendix:prompt_direct}

The prompt in Table~\ref{tab:prompt_direct} is used in both Direct and the fact triplet verification step of GraphCheck to assess whether a claim is true or false, given the retrieved evidence. 

\begin{table}[h]
    \centering
    \renewcommand{\arraystretch}{1.2}
    \small
    \begin{tabular}{|p{0.95\textwidth}|}
    \hline
    \vspace{0.01em}
    \textit{Evidence:} \texttt{\{evidence\}} \\
    \textit{Claim:} \texttt{\{claim\}} \\
    Is the claim true or false? \\
    \textit{Answer:} \\ [0.5em]
    \hline
    \end{tabular}
    \caption{Prompt used for Direct and fact triplet verification.}
    \label{tab:prompt_direct}
\end{table}

In the case of Direct, \texttt{\small \{claim\}} refers to the original input claim.
In the fact triplet verification step of GraphCheck, it refers to a triplet converted into a natural language sentence (i.e., with \texttt{\small [SEP]} tokens removed). \texttt{\small \{evidence\}} denotes either (i) the concatenation of top-$k$ documents retrieved using the original claim or the converted triplet as the query, or (ii) each individual document within the top-$k$.


\subsection{Strategy Selection}
\label{appendix:prompt_strategy_selection}

The prompt in Table~\ref{tab:prompt_strategy_selection} is used for strategy selection—deciding whether to use Direct or more systematic GraphCheck to verify a given claim. Specifically, it asks if the retrieved evidence documents contain sufficient information to support or refute the claim. If the answer is ``yes'' (sufficient), the claim is verified using Direct. If the answer is ``no'' (insufficient), the claim is verified using GraphCheck.

\begin{table}[h]
    \centering
    \renewcommand{\arraystretch}{1.2}
    \small
    \begin{tabular}{|p{0.95\textwidth}|}
    \hline
    \vspace{0.01em}
    \textit{Evidence:} \texttt{\{evidence\}}\\
    \textit{Claim:} \texttt{\{claim\}}\\
    Does the evidence contain sufficient information to support or refute the claim? Yes or no?\\
    \textit{Answer:} \\[0.5em]
    \hline
    \end{tabular}
    \caption{Prompt used for strategy selection.}
    \label{tab:prompt_strategy_selection}
\end{table}

Here, \texttt{\small \{claim\}} refers to the original input claim.
\texttt{\small \{evidence\}} is the concatenation of the top-$k$ documents retrieved using the original claim as the retrieval query.

\section{Examples of Graph Construction Error}
\label{appendix:graph_error_examples}

\begin{table}[h]
    \centering
    \renewcommand{\arraystretch}{1.2}
    \small
    \begin{tabular}{p{0.95\textwidth}}
    \hline
    \vspace{0.01em}
    \textbf{(1) Hallucinated Latent Entities}

    \vspace{0.5em}
    \quad \textit{\# Claim:} \\
    \quad Age of Chance and Vector were both a rock band \\
    \quad \textit{\# Latent Entities:} \\
    \quad (ENT1) [SEP] is [SEP] a rock band \\
    \quad (ENT2) [SEP] is [SEP] a rock band \\
    \quad \textit{\# Triples:} \\
    \quad Age of Chance [SEP] is [SEP] (ENT1) \\
    \quad Vector [SEP] is [SEP] (ENT2)

    \vspace{0.5em}
    \quad \textbf{Issue:} Non-latent entities (\textit{``Age of Chance'', ``Vector''}) are mistakenly marked as latent, leading to unnecessary and potentially erroneous infilling. \\

    \vspace{1em}
    \textbf{(2) Decomposition Error}

    \vspace{0.5em}
    \quad \textit{\# Claim:} \\
    \quad The place where MLB team owner Doc Watson (baseball) was born is named after the only Catholic signatory to the Declaration of Independence, Charles Carroll of Carrollton. \\
    \quad \textit{\# Latent Entities:} \\
    \quad (ENT1) [SEP] is [SEP] a place \\
    \quad \textit{\# Triples:} \\
    \quad Doc Watson (baseball) [SEP] was born in [SEP] (ENT1) \\
    \quad (ENT1) [SEP] is named after [SEP] Charles Carroll of Carrollton \\
    \quad Charles Carroll of Carrollton [SEP] is [SEP] the only Catholic signatory to the Declaration of Independence

    \vspace{0.5em}
    \quad \textbf{Issue:} Important context (\textit{``MLB team owner''}) is dropped during decomposition. \\\\

    \hline
    \end{tabular}
    \caption{Examples of graph construction error in GraphCheck.}
    \label{tab:graph_error_examples}
\end{table}

\end{document}